%% file: main.tex
\documentclass[11pt]{article}

\usepackage{acl}

\input{style}
\input{math_commands}
\usepackage{times}
\usepackage{latexsym}

\usepackage[T1]{fontenc}

\usepackage[utf8]{inputenc}

\usepackage{microtype}

\usepackage{inconsolata}

\usepackage{graphicx}

\title{Mixture-of-Personas Language Models for Population Simulation}

\author{
 \textbf{Ngoc Bui\textsuperscript{1}},
 \textbf{Hieu Trung Nguyen\textsuperscript{2}},
 \textbf{Shantanu Kumar\textsuperscript{1}},
\\
 \textbf{Julian Theodore\textsuperscript{1}},
 \textbf{Weikang Qiu\textsuperscript{1}},
 \textbf{Viet Anh Nguyen\textsuperscript{2}},
 \textbf{Rex Ying\textsuperscript{1}},
\\
 \textsuperscript{1}Yale University,
 \textsuperscript{2}The Chinese University of Hong Kong,
\\
\small{\texttt{\{ngoc.bui, shantanu.kumar, jt2386, weikang.qiu, rex.ying\}@yale.edu, \{thnguyen, nguyen\}@se.cuhk.edu.hk}}
}
\begin{document}

\maketitle
\begin{abstract}
  Advances in Large Language Models (LLMs) paved the way for their emerging applications in various domains, such as human behavior simulations, where LLMs could augment human-generated data in social science research and machine learning model training. However, pretrained LLMs often fail to capture the behavioral diversity of target populations due to the inherent variability across individuals and groups. To address this, we propose \textit{Mixture of Personas} (MoP), a \textit{probabilistic} prompting method that aligns the LLM responses with the target population. MoP is a contextual mixture model, where each component is an LM agent characterized by a persona and an exemplar representing subpopulation behaviors. The persona and exemplar are randomly chosen according to the learned mixing weights to elicit diverse LLM responses during simulation. MoP is flexible, requires no model finetuning, and is transferable across base models. Experiments for synthetic data generation show that MoP outperforms competing methods in alignment and diversity metrics.
\end{abstract}

\section{Introduction}

\begin{figure}
    \centering
    \includegraphics[width=\linewidth]{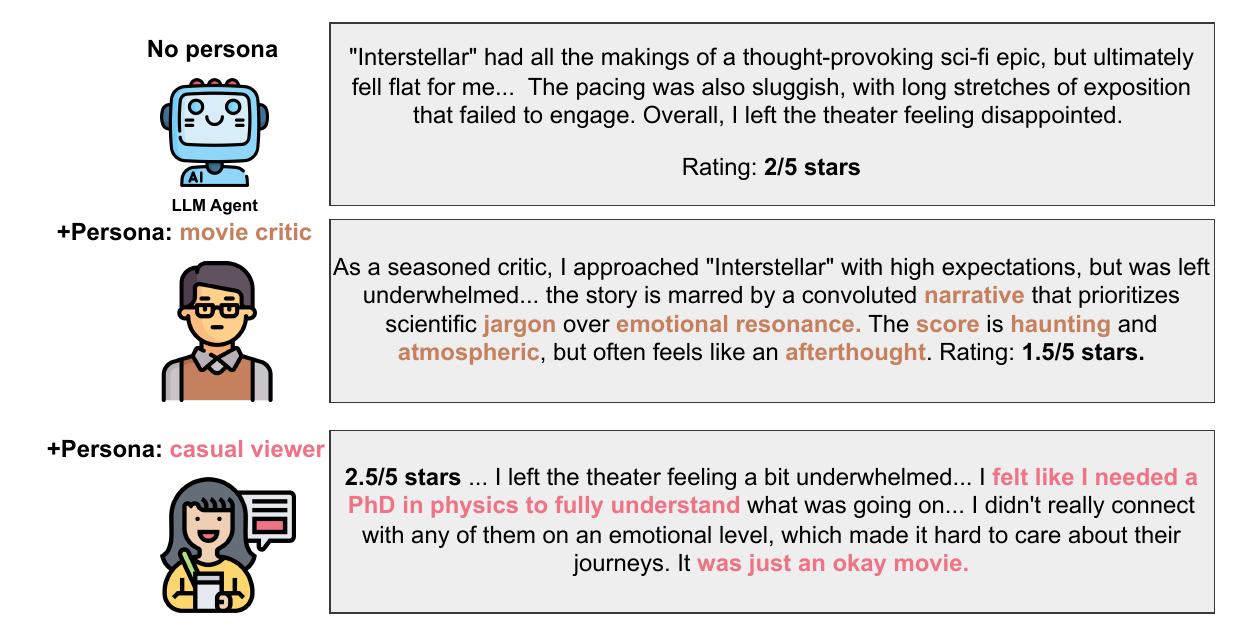}
    \caption{Sampling from foundational LLM agents frequently yields repetitive and generic responses. Meanwhile, prompting with personas can create more tailored, specific responses. The highlighted words in the figure correspond to the prompted personas.}
    \label{fig:persona_prompt}
\end{figure}

The impressive capability of Large Language Models (LLMs) to generate human-like output has enabled their application across various domains, where their responses can complement or substitute human-generated data, providing a scalable approach to address data limitations~\citep{argyle2023out}. Prominent examples include simulating human behaviors in social science~\citep{aher2023using, argyle2023out}, modeling economic agents and decision-making in economics~\citep{horton2023large}, analyzing political trends and electoral dynamics in political science~\citep{bisbee2023synthetic}, or generating synthetic training data~\citep{li2023synthetic, tornberg2023chatgpt}. Natural human responses in those applications often reflect diverse behaviors or preferences of different personas shaped by demographic, cultural, and societal variations of the target population~\cite {zhao2023group}. Modeling the distribution of those behaviors is crucial for generating realistic and contextually relevant outputs~\citep{sorensen2024roadmap}. A common approach to achieving tailored responses is to prompt LLMs with a persona that simulates a specific group's behavior, language, and preferences; see Figure~\ref{fig:persona_prompt}. However, recent studies show that LLM's responses often lack diversity and exhibit significant biases \citep{yu2024large}, and this downside persists even when LLMs are prompted with a persona \citep{santurkar2023whose}. 

Some efforts has been made toward improving the steerability LLMs, enabling them to produce outputs aligned with specific user intentions or personas. Approaches addressing steerability generally fall into two main categories: prompt engineering~\citep{hwang2023aligning, chen2024large} and fine-tuning using personalized datasets~\citep{li2023steerability, sun2024persona, zhao2023group, choi2024picle}. Although these methods aim to capture the behaviors, preferences, and communication styles of particular user groups, both techniques encounter distinct challenges. Prompt engineering tailored for each user is both intricate and resource-intensive. Meanwhile, learning-based methods require access to personal data, which is often scarce or expensive. The reliance on personal data also raises privacy concerns, restricting the practical applicability of these approaches. 

Furthermore, diversity sampling remains a notable challenge in LLMs, especially when simulating responses representative of diverse populations. Current methods~\citep{choi2024picle} typically rely on a fixed, optimal selection of few-shot examples reused across multiple downstream tasks, complemented primarily by temperature scaling to enhance response diversity. However, relying solely on temperature scaling can be inadequate for generating semantically diverse outputs, frequently resulting in a trade-off between quality and diversity or causing outputs to collapse into semantically similar responses~\citep{chang2023kl}.

\noindent\textbf{Proposed work.} We address the steerability problem in LLMs, focusing on aligning model responses with the characteristics of a target population. To achieve this, we propose the \textit{Mixture of Persona} (MoP), a probabilistic prompting framework that leverages persona descriptions and in-context exemplars to steer responses. MoP functions as a contextual mixture model comprising multiple LLM agents, each characterized by a persona prompt that is either user-defined or synthesized from observed target population responses. During response generation, personas are probabilistically selected based on mixing weights, enabling the model to produce customized and contextually relevant outputs.

Since the naive MoP approach described above may still be susceptible to biases and limited response diversity, we enhance it by incorporating in-context examples to better align LLM responses with population characteristics. These in-context exemplars are drawn anonymously from a representative pool of the target population, guided by learnable weights. Importantly, our method does not require direct associations between individual personas and specific examples, thus minimizing reliance on personal data. Instead, each exemplar's influence on a persona is controlled through a secondary set of mixing weights, forming a two-level hierarchical mixture model. The first level manages persona selection, while the second level determines exemplar weighting. This hierarchical structure enables MoP to effectively represent the diversity and complexity inherent in population-level behaviors, simultaneously mitigating biases in LLM-generated responses.

We conduct extensive experiments to evaluate the effectiveness of MoP against existing prompting methods. Specifically, our experiments span two main scenarios: (1) simulating human-like opinions in tasks such as generating movie and restaurant reviews as well as news articles, and (2) creating synthetic data for downstream classification tasks. The results demonstrate that MoP significantly enhances alignment with target population responses, achieving a 58\% improvement in FID scores and a 28\% increase in MAUVE scores compared to the strongest baseline. Additionally, MoP generates more diverse responses, effectively capturing nuanced variations in population-level behavior without compromising response quality. We further demonstrate the transferability of MoP: a model trained on Llama3-8B-Instruct can directly generalize to other base models such as Gemma-9B-Instruct and Mistral-7B-Instruct in a plug-and-play fashion, eliminating the need for retraining.

\section{Problem Setting}

\begin{figure*}
    \centering
    \includegraphics[width=\linewidth]{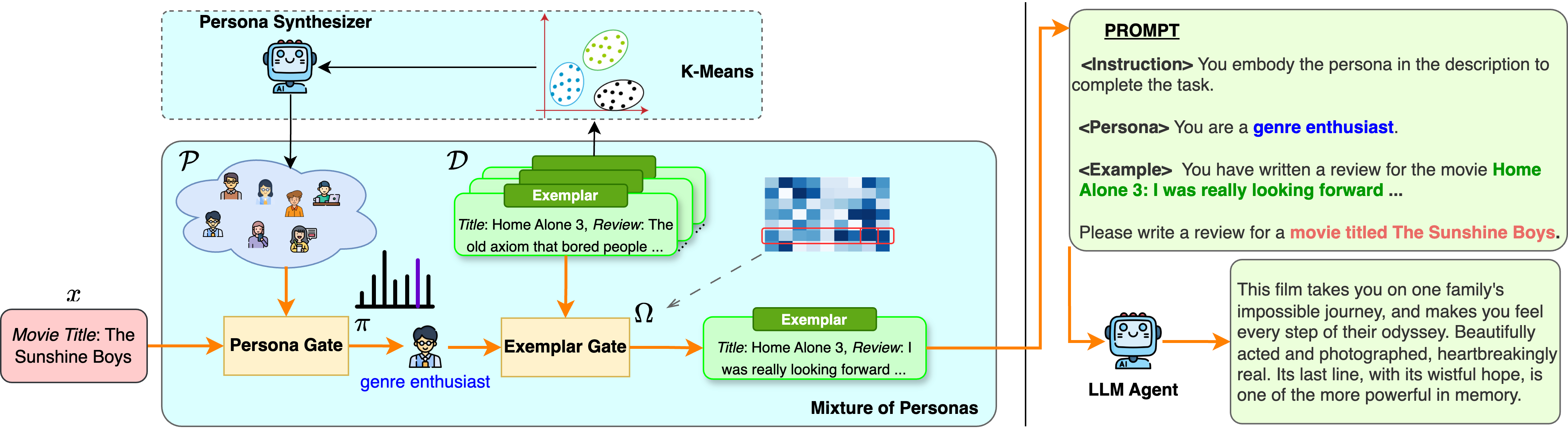}
    \caption{The generation pipeline for the Exemplar-based Mixture of Personas (MoP) operates as follows: Given a movie review $x$, MoP first samples a persona based on the learnable mixing weight $\pi$. Next, MoP selects an exemplar randomly from the observation pool according to the mixing weight $\Omega$. The selected persona and exemplar are then concatenated with the input context to create a personalized prompt used to sample from a base LLM agent. The dashed block indicates the process of persona synthesis.}
    \label{fig:pipeline}
\end{figure*}

We consider the problem of simulating human behaviors or preferences at the population level using prior knowledge from pre-trained LLMs~\citep{sorensen2024roadmap}. Let $\mc P$ be a population composed of $K$ groups of interest. Each group $k \in \{1, \ldots, K\}$ can be characterized by a persona $g_k$ that reflects the traits and motivations driving the behaviors of individuals within that group. Let $\mc D = \{(x_i, y_i)\}_N$ be a recorded data set from the population of interest $\mc P$. For synthetic data generation tasks such as movie review generation, $x_i$ could be an input context (\textit{e.g.}, movie title), and $y_i$ could be a human response (\textit{e.g.}, a movie review). Note that a given input $x$ could be associated with multiple, diverse responses $y$ produced by different individuals in the population $\mc P$ with different preferences. 

In real-world applications, the persona description $\{g_k\}_K$ can be pre-defined by the users according to the task at hand or automatically synthesized from the record set $\mc D$. We will consider both variants in the experiments. It is worth noting further that, different from ~\citet{zhao2023group} and~\citet{choi2024picle}, we do not require access to preference or personal data generated by persona $g_k$ (\textit{e.g.,} $g_k-(x_i, y_i)$ pairs); therefore, our setting could be considered as an `unsupervised' setting of the steerability problem of LLMs.

\noindent\textbf{Notations.} In what follows, we use $p_{LM}(y | x)$ to denote the probability density of the pre-trained LLM for an output $y$ given the input $x$ as a prompt and $p_{LM} (y | g_k, x)$ to denote the probability given the concatenated input $[g_k, x]$, in that order.

\section{Methodology}\label{sec:method}

In this section, we propose the \textit{Mixture of Personas} (MoP), a contextual mixture model of LM agents designed to simulate the diverse preferences of individuals within a target population $\mc P$. In MoP, each agent is characterized by a persona prompt that encapsulates a specific subgroup's aggregated behaviors or preferences. Furthermore, the agent's responses are guided by an exemplar, randomly selected from the anonymous data set $\mc D$ based on a learnable mixing weight, making MoP a two-level hierarchical mixture of LLM agents. The overall pipeline is depicted in Figure~\ref{fig:pipeline}. For clarity, we assume that the persona descriptions $\{g_k\}_K$ are pre-defined, postponing the discussion of persona synthesis to the end of this section.

\subsection{Mixture of Personas}\label{sec:mop}

We propose modeling the population's responses as a mixture of responses generated by $K$ constituent personas. Specifically, given an input context $x$, the probability of receiving a response $y$ from the population $\mc P$ can be decomposed into group-based probabilities as
\begin{equation}
    p(y | x) = \sum_{k = 1}^K \pi_k p_{LM} (y | g_k, x) ~~\text{with}~~\sum_k \pi_k = 1,
\end{equation}
where $g_k$ is a persona prompt representative for group characteristics $k$ and $\pi_k \in [0, 1]$ is the mixing weight specifying the group's propensity. Since population members may contribute differently to different input contexts\footnote{For instance, in a movie review scenario, an art house or indie film may receive more reviews from critics than casual viewers, whereas an action movie might attract more reviews from casual viewers than critics.},
we parametrize these mixing weights using a simple gating network that is conditionally dependent on the input context $x$, following~\citet{jordan1994hierarchical}.

Specifically, we leverage a pre-trained sentence encoder to capture the semantics of the input context $x$ and persona prompt $g_k$
\begin{align} \label{eq:emb}
    \mbf x = \mbf W_x h(x), \quad \mbf g_k = \mbf W_g h(g_k).
\end{align}
Here, $h(\cdot) \in \R^{d'}$ is a pre-trained sentence encoder and $\mbf W_x, \mbf W_g \in \R^{d' \times d}$ are learnable parameters of the gating network. We now define the \textit{persona gate} $\pi$ based on the similarities between the input context and the persona prompts
\begin{align}
    \pi = \softmax(\mbf x^\top \mbf g_1, \ldots, \mbf x^\top \mbf g_K) \in [0, 1]^K,
\end{align}
where the softmax normalizes the vector of $K$ logits into probability mixing weights that determines the probability of selecting persona $g_k$ given the input context $x$.
 
This persona prompt allows us to leverage abundant prior knowledge of LLMs to steer its responses toward the behaviors of the group $g_k$~\citep{wang2023unleashing, chen2024large, tseng2024two}. However, this direct prompting technique alone is often insufficient, as LLMs are generally pre-trained on massive, global datasets, which might not be aligned with our targeted group~\citep{zhao2023group}. Additionally, it is known that relying solely on temperature scaling for LLM decoding may not effectively generate semantically diverse responses~\citep{chang2023kl}. As a result, simulated responses tend to collapse into similar outputs. In the next section, we propose a method to address the above problems to elicit diverse responses from the pre-trained LLMs and align them to the group's preference using exemplars in the record set $\mc D$.

\subsection{Exemplar-based Mixture of Personas}\label{sec:exemplar_mop}
To address the aforementioned problem, we propose to combine the persona prompt with an exemplar, acting as historical data to guide the LLM agent and eliciting more diverse responses. Specifically, given an anonymous dataset $\mc D = \{(x_i, y_i)\}_N$, we model the distribution $p_{LM}( \cdot | g_k)$ using another layer of the mixture model that aggregates the responses of the LLM agent $k$, augmented with varying exemplars
\begin{align}\label{eq:likelihood}
    &p(y | x, \mc D) = \sum_{k = 1}^K \pi_k \sum_{j=1}^N \Omega_{kj} p^{\tau_k}_{LM} (y | g_k, x_j, y_j, x), \\
    &\text{where}~\sum_k \pi = 1 \quad \text{and} \quad \sum_j \Omega_{kj} = 1 ~~\forall k. \notag
\end{align}
Above, $\Omega_{kj}$ denotes the importance weight of the exemplar $(x_j, y_j)$ to the subpopulation $k$ and $p(y | g_k, x_j, y_j, x)$ represents the probability that an individual in group $k$ will respond to the context $x$ with $y$, given that they previously responded to context $x_j$ with $y_j$. 

Similar to the persona gate, we parameterize exemplar selection with a simple \textit{exemplar gate $\Omega$}, which selects an exemplar based on the input context $x$ and the chosen persona $g_k$. Specifically, we define the exemplar gate $\Omega_{k:}$ corresponding to persona $k$ as follows:
\begin{equation*}
\resizebox{\linewidth}{!}{
    $\Omega_{k:} = \softmax(\mbf x^\top \mbf e_1 + \mbf g_k^\top \mbf e_1, ..., \mbf x^\top \mbf e_N + \mbf g_k^\top \mbf e_N),$
}
\end{equation*}
where $\mbf e_i = \mbf W_e h(e_i)$ is the embedding of the exemplar $i$. The input context $\mbf x$ and persona prompt $\mbf g_k$ are computed as defined in Eq.~\eqref{eq:emb}. Here, the utility of an exemplar explicitly depends on both input context $x$ and the persona prompt $g_k$. The softmax function is applied along each column of $\Omega$ to ensure that the mixing weights are summed to one, \textit{i.e.}, $\sum_j \Omega_{kj} = 1$ for all $j$. Additionally, we add a learnable temperature parameter $\tau_k$ for each persona to normalize LLM output logits, thereby controlling the diversity of responses for each persona. This parameter will be the temperature hyperparameter for the LLM's sampling algorithm during simulation. In summary, MoP is an instance of the two-level hierarchical mixture-of-experts model~\citep{jordan1994hierarchical}, where each expert is an LLM agent prompted with a persona and an exemplar. 

This prompting pipeline can be viewed as an in-context learning method for augmenting LLM instructions, a strategy widely applied across various LLM tasks~\citep{brown2020language}. However, our method introduces two key differences from existing approaches. \textit{First}, we operate in an \textit{unsupervised} setting, where no explicit personal data pairs $\{g_k, (x_i, y_i)\}$ are required. Instead, the relevance of exemplars to the persona $g_k$ is learned and controlled by the mixing weight $\Omega$. \textit{Second}, while existing methods typically select a fixed, optimal set of few-shot examples to be reused across different instances in downstream tasks~\citep{choi2024picle}, our approach randomly selects exemplars based on the weight $\Omega$, which enhances the diversity of simulated responses.

\subsection{Optimizing The Gating Networks}
We optimize the gating network to fit the population observation $\mc D$ by treating MoP learning as a maximum log-likelihood of the population observations $\mc D$. Let us denote $\theta$ as the gating network's learnable parameters. The log-likelihood of the observation $\mc D$ will be computed by
\begin{align}\label{eq:loglikelihood}
&\log (\theta; \mc D) = \\
\nonumber
&\sum_{i}^N \log \left( \sum_{k}^K \sum_{j}^N \pi_k \Omega_{kj} p^{\tau_k}_{LM}(y_i | g_k, x_j, y_j, x_i) \right).
\end{align}
The above computation requires $K \times N$ LLM forward passes to compute the log-likelihood estimate, which is prohibitively expensive, especially when $N$ and $K$ are large. To address this problem, we adopt a sparsely gating mechanism widely used in the mixture-of-experts (MoEs) literature~\cite {shazeer2017outrageously}. For any context $x$, we only estimate the log-probability of the pre-trained LLM for top-$M$ pairs of persona and exemplars with the highest probability $\pi_k\Omega_{kj}$. This allows our training to scale up to thousands of exemplars and personas. Further, notice that we use the same observation set $\mc D$ as the exemplars for steering LLM output towards observations in $\mc D$. This can lead to over-optimization of the mixture model since, when estimating the log-likelihood $p(y_i | x_i, \mc D)$, the observation $(x_i, y_i)$ is already seen in the dataset $\mc D$~\citep{mallapragada2010non}. To address this problem, we randomly mask the target example during training to avoid using the same exemplar to predict itself.

It is worth noting further that we exclusively train the gating network while keeping the pre-trained LLM entirely fixed, requiring access only to the LLM's output logits. As a result, our gating mechanism is transferable and can be applied to other pre-trained LLMs in a plug-and-play fashion without the need for retraining.

\subsection{Population Simulation}
Similar to other mixture models, our model admits an equivalent representation using the following generative process
\begin{align*}
    c | x &\sim \mathrm{Cat}(\pi) & \forall c \in [1, K], \\
    h | c, x &\sim \mathrm{Cat}(\Omega_{k:}) & \forall h \in [1, N], \\
    y | h, c, x &\sim p^{\tau_k}_{LM}(y | g_c, y_h, x),
\end{align*}
where $c$ and $h$ are latent variables indicating the selected group and exemplar for the simulated records. To simulate responses from the population, we first sample $c$ and $h$ sequentially from two categorical distributions and then sample responses from the LLM agent using the corresponding persona and exemplar. This sampling process is memory and computation efficient because it does not incur an overhead by switching between different personalized models like existing finetuning approaches~\citep{yu2024neeko}. 

\subsection{Persona Synthesis}

Although persona descriptions $\{g_k\}_K$ can be predefined by users depending on the task at hand, there are scenarios where it is preferable to synthesize personas directly from the given dataset $\mc D$. Given the assumption that no personal records are available, we utilize a pre-trained sentence encoder and LLM to cluster and generate the persona descriptions. Specifically, we first encode all records into a shared embedding space using the `all-mpnet-base-v2` sentence encoder and then apply the $K$-means algorithm to split the records set into $K$ clusters. For each cluster, we prompt a pre-trained LLM to summarize the records within the cluster, producing a persona description for the group. Details of the prompt used to generate persona descriptions can be found in Appendix~\ref{sec:prompts}.

\section{Experiments}
In this section, we conduct extensive experiments to investigate the effectiveness of the proposed methods, aiming to answer the following three research questions:
\begin{enumerate}[leftmargin=12mm,start=1,label={\bfseries RQ\arabic*:}]
    \itemsep0em
    \item Can MoP steer the output of LLMs towards the population of interest?
    \item Can we utilize the MoP prompting technique to generate high-quality data for training ML models on task-specific applications? 
    \item Can MoP be transferable to different pre-trained LLMs in a plug-and-play manner?
\end{enumerate}
To answer the above research questions, we conduct two main experiments:
\noindent\textit{Steerability}: We generate news articles and movie/restaurant reviews to demonstrate that a pre-trained LLM with MoP prompting can simulate the distribution of human-generated content while maintaining its diversity.
\noindent\textit{Synthetic Data for Training ML Models}: We show that adapting MoP prompting can create high-quality training samples for classification tasks, such as topic or sentiment classification.

\begin{table*}[ht]
    \caption{Alignment and diversity of the generated datasets using our MoP with respect to the golden test set. The $*$ symbol indicates datasets obtained directly from the authors' released datasets~\citep{yu2024large}.} 
    \label{tab:steerability}
    \resizebox{\textwidth}{!}{
    \begin{filecontents*}{quality_diversity.csv}
,AgNews,AgNews,AgNews,Yelp,Yelp,Yelp,sst2,sst2,sst2,imdb,imdb,imdb
,FID$\downarrow$,MAUVE$\uparrow$,KL Cosine $\downarrow$,FID$\downarrow$,MAUVE$\uparrow$,KL Cosine $\downarrow$,FID$\downarrow$,MAUVE$\uparrow$,KL Cosine $\downarrow$,FID$\downarrow$,MAUVE$\uparrow$,KL Cosine $\downarrow$
ZeroGen (GPT4)*,3.248,0.553,0.506,3.427,0.517,1.828,3.525,0.559,0.911,3.737,0.518,0.200
AttrPrompt (GPT4)*,2.992,0.585,0.399,2.599,0.570,0.428,4.133,0.550,1.834,2.477,0.566,0.789
ZeroGen,3.535,0.587,0.241,1.888,0.682,0.173,3.965,0.550,0.800,3.529,0.537,0.195
PICLe,2.200,0.740,0.490,1.769,0.702,0.265,3.531,0.562,3.180,2.870,0.609,1.459
ProGen,1.980,0.767,0.103,2.975,0.586,1.332,4.736,0.615,2.023,3.305,0.612,1.045
AttrPrompt,2.193,0.648,0.150,1.816,0.651,0.143,3.878,0.555,1.393,2.505,0.549,0.492
\textbf{MoP},\textbf{0.951},\textbf{0.871},\textbf{0.069},\textbf{0.948},\textbf{0.826},\textbf{0.067},\textbf{1.131},\textbf{0.855},\textbf{0.319},\textbf{0.771},\textbf{0.865},\textbf{0.039}
Improvement ($\%$),51.970,13.559,33.010,46.410,17.664,53.147,67.969,39.024,60.125,68.874,41.340,80.000
\end{filecontents*}
\pgfplotstabletypeset[
        col sep=comma,
        string type,
        every head row/.style={before row=\toprule},
        every row no 0/.style={after row=\midrule},
        every head row/.style={output empty row, before row={%
                \toprule \multirow{3}{*}{Method} &
                \multicolumn{3}{c}{AgNews} & 
                \multicolumn{3}{c}{Yelp}&
                \multicolumn{3}{c}{SST-2}&
                \multicolumn{3}{c}{IMDB}\\ 
                \cmidrule(r){2-4} \cmidrule(r){5-7} \cmidrule(r){8-10} \cmidrule(r){11-13}
        }},
        every last row/.style={after row=\bottomrule},
        every last row/.style={before row=\midrule, after row=\bottomrule},
   ]{quality_diversity.csv}
    }
    \vspace{-2mm}
\end{table*}

\begin{table}[ht]
    \centering
    \caption{Performance on downstream classification task using generated datasets to train a sentiment classifier (Yelp, SST-2, and IMDB) or news topic classifier (AgNews). F1 scores are calculated on the golden test set.}
    \label{tab:f1score}
    \small
    \begin{tabular}{ccccc}
        \toprule
         & AgNews & Yelp & SST-2 & IMDB\\ \toprule
        Golden data   & 0.903 &	0.896 &	0.919 &	0.877   \\
        \hdashline
        Zerogen   & 0.624&	0.860 &	0.766 &	0.821   \\ 
        ProGen   & 0.722 & 0.843 & 0.785 & 0.810  \\
        PICLe   & 0.759 & 0.738 & 0.833 & 0.815  \\
        AttrPrompt   & 0.836 & 0.864 & 0.838 & 0.793  \\
        \textbf{MoP} & \textbf{0.871} & \textbf{0.867} & \textbf{0.845} & \textbf{0.865} \\ 
        Improvement (\%) & 4.190 & 0.347 & 0.835 & 5.359 \\ 
        \bottomrule
        \end{tabular}
    \label{tab:downstream_task}
    \vspace{-6mm}
\end{table}

\subsection{Setup}
We describe the datasets, baselines, and metrics used throughout the experiment section.

\noindent\textbf{Datasets.} We employ four datasets that are commonly used in dataset generation literature~\citep{yu2024large}, including \textit{AGNews}~\cite{zhang2015character}, \textit{Yelp Reviews}~\cite{zhang2015character}, \textit{SST-2 (Stanford Sentiment Treebank)}~\cite{socher-etal-2013-recursive}, and \textit{IMDB Reviews}~\cite{maas-etal-2011-learning}. AGNews includes news articles from AG News; each article belongs to one of four topics: `world', `sports', `business', and `technology'. Yelp is a restaurant review dataset while SST-2 and IMDB are movie review datasets; each review has a binary label indicating positive or negative review. Note that we use labels for testing purposes only. Each dataset includes two splits for training and testing. We train MoP on the training dataset and evaluate the synthetic data with the test set. Similar to~\cite{yu2024large}, we call the test set for testing a golden dataset. More details and statistics of the datasets are provided in Appendix~\ref{sec:add_exp_setting}.

\textbf{Baselines.} We compare our method to prompting baselines in dataset generation literature: \textit{ZeroGen}~\cite{ye2022zerogen}, \textit{AttrPrompt}~\cite{yu2024large}, \textit{ProGen}~\cite{ye2022progen} and steerability literature: \textit{PICLe}~\cite{choi2024picle}. Here, ZeroGen is a simple zero-shot, context-dependent prompting method, and AttrPrompt is an attribute-randomized prompting method to increase the diversity of synthesized samples. ProGen uses the influence function to weigh and choose synthesized samples as in-context examples. PCILe is an in-context prompting method to address the steerability problem of LLMs, where in-context examples are chosen according to the weight given by the logit difference between the persona-finetuned and the pretrained models. Appendix~\ref{sec:add_exp_setting} provides more details of the baselines. In the following, we refer to MoP as our Exemplar-based Mixture of Personas described in Section~\ref{sec:exemplar_mop}.

\textbf{Metrics.} We follow~\citet{pillutla2021mauve} and \citet{yu2024large} to evaluate the alignment and diversity of the simulated data in the embedding space. Throughout, we use a pre-trained sentence encoder `all-mpnet-base-v2'\footnote{https://huggingface.co/sentence-transformers/all-mpnet-base-v2} from~\citet{song2020mpnet} as a text encoder to map generated sentences into a common vector space. We then compute \textit{FID} (Fréchet Inception Distance)~\cite{heusel2017gans} and \textit{MAUVE}~\cite{pillutla2021mauve, pillutla2023mauve} to evaluate the alignment between the simulated responses and the golden dataset. To evaluate the diversity of generated responses compared to the golden dataset, we report the KL-divergence of two histograms constructed by pairwise cosine similarity values of the generated responses and the golden dataset. We call this diversity metric KL Cosine. Note that MAUVE can also capture the diversity alignment to some extent~\cite{pillutla2021mauve}. More details for the metrics are provided in Appendix~\ref{sec:add_exp_setting}.


\textbf{Implementation Details\footnote{The source code for our implementation is released at~\url{https://github.com/ngocbh/MoP}}.}
For all experiments, we use the same Llama3-8B-Instruct~\cite{dubey2024llama} as the base model for our method and other baselines. For MoP implementation, we choose the number of personas to be 100 and randomly select 1,000 observations in the training dataset $\mc D$ as the set of exemplars. We then run $K$-Means and the persona synthesizer to extract 100 persona descriptions. After that, the personas and exemplars will be fixed during training and inference. Throughout, we use `all-mpnet-base-v2' as the sentence encoder $h(\cdot)$ and set the hidden dimension for linear layers in the gating network as 128. During training, we choose top $M = 4$ persona-examples pairs for each input context $x$ for LLM forwards. We initially set the temperature $\tau = 0.6$ and let it be learnable during the training along with our gating networks. We use the default temperature $\tau = 1$ for other baselines as discussed in~\citet{yu2024large} and \citet{ye2022zerogen}. Other generation hyperparameters of the base LLM model are set by default. For each method, we generate 5,000 synthetic responses and measure the evaluated metrics compared to the golden test set.

\subsection{Steerability (RQ1)}\label{sec:exp_steer}
Table~\ref{tab:steerability} shows that MoP can significantly outperform other baseline prompting methods. Specifically, our method can outperform the best baseline by 58.8\% in terms of FID and by 27.9\% in terms of MAUVE, averaged over all datasets. This result demonstrates that MoP can effectively steer LLM's outputs to the target responses of the target population. Besides the improvement in alignment scores, KL Cosine metrics also indicate that our responses are more diverse than other baselines. 
\begin{figure*}
    \centering
    \includegraphics[width=\linewidth]{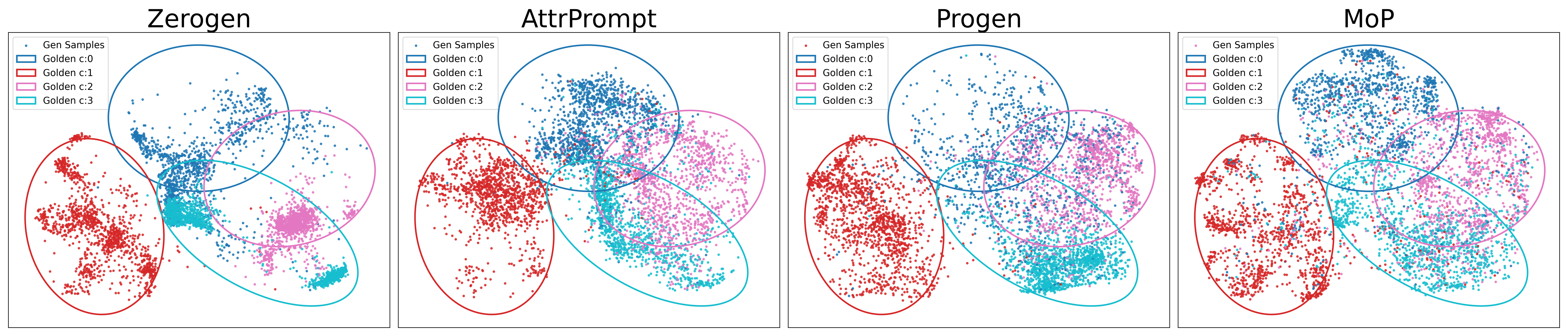}
    \caption{Diversity comparisons on Agnews of different prompting methods. The embeddings are computed using `all-mpnet-base-v2'. The scatted points are synthesized samples with the colors indicating the corresponding labels. The circle lines indicate 2-std confidence ellipses of the golden test set. It can be seen that MoP offers synthesized samples that are diverse and aligned with the golden test set.}
    \label{fig:umap}
\end{figure*}

\begin{figure}
    \centering
    \includegraphics[width=\linewidth]{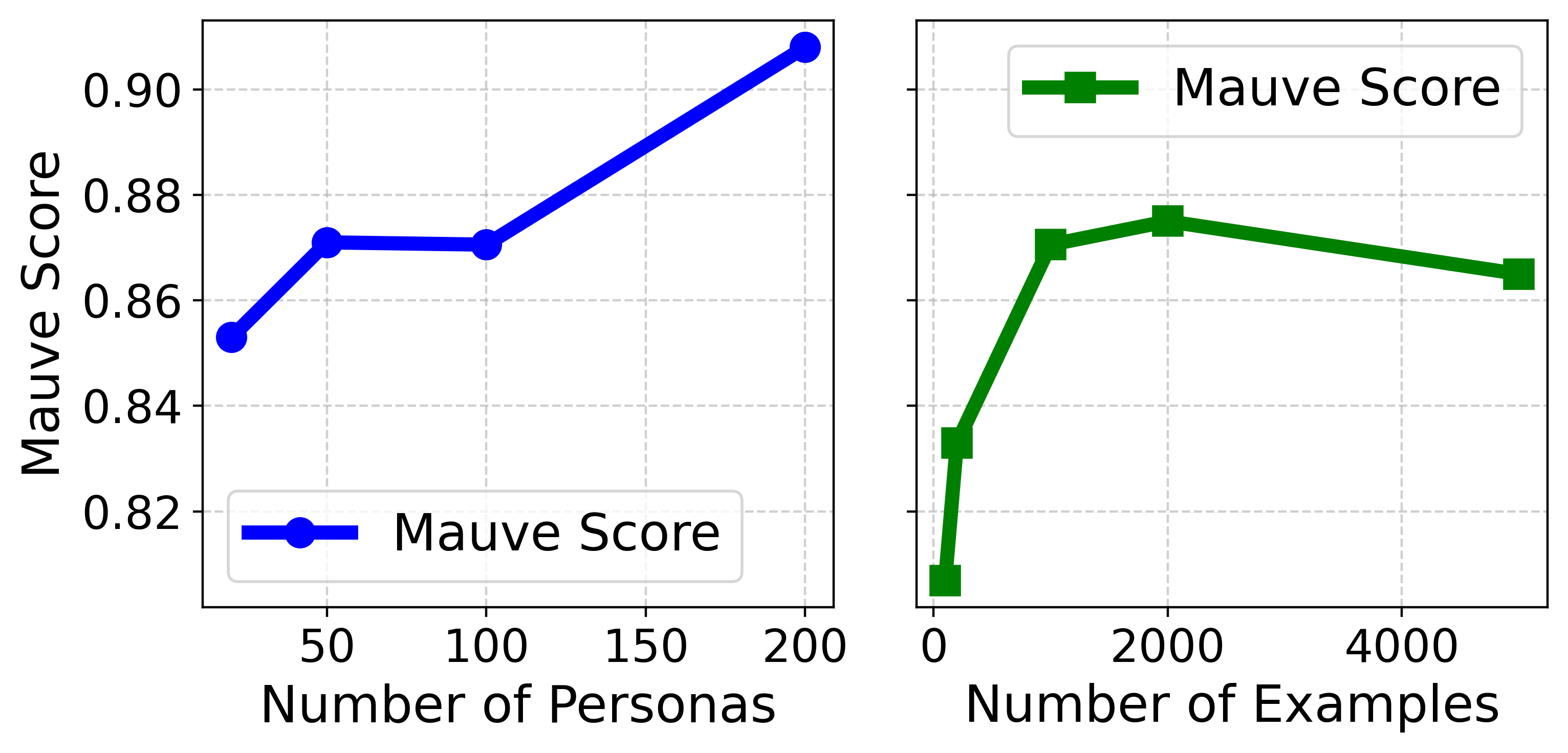}
    \vspace{-3mm}
    \caption{Mauve scores with varying the number of personas and number of examples.}
    \label{fig:abl_vary}
    \vspace{-6mm}
\end{figure}
\subsection{Synthetic Data Generation (RQ2)}
We explore the two methods for leveraging the trained MoP in Section~\ref{sec:exp_steer} to generate synthetic data tailored to task-specific classification tasks. One is applied for topic classification (Agnews) and the other is applied to sentiment classification tasks (Yelp, SST2, and IMDB).

For Agnews, we label the personas mined by the persona synthesizer and use it as the label for the sample generated by the chosen persona. We classify the persona description into four categories: world reporter, sports reporter, business reporter, and technology reporter. For completeness of the pipeline, we prompt pretrained LLM to do this simple task. The prompt is provided in Appendix~\ref{sec:prompts}.

For the second method, we augment the input prompt with the following context for generating 2,500 positive reviews (similarly for negative):

\noindent\textit{``You watched the movie \{title\} and had positive impression. Please write a review for the movie:''}

We train a DistilBERT model~\citep{sanh2019distilbert} using 5,000 synthesized samples and evaluate its performance on the golden dataset, reporting the F1-score in Table~\ref{tab:downstream_task}. Results show that the synthetic dataset generated by Llama3 with MoP prompting achieves up to a 2.68\% improvement over the AttrPrompt baseline. The smaller performance gap observed with the Yelp dataset likely occurs because Llama3-8B-Instruct already aligns closely with the Yelp test distribution. This can be seen by the performance difference between ZeroGen and the golden training data is comparatively smaller than in other datasets, leaving less room for further improvement.

To further illustrate the advantage of MoP prompting, we plot the embedding space provided in Figure~\ref{fig:umap}. The result indicates that our method better matches the diversity of the golden datasets compared to other baselines. Meanwhile, simulated responses of other baselines often collapse or cannot cover the diversity of the true responses.

\subsection{Transferability (RQ3)}
In this section, we evaluate the transferability of MoP to new model architectures. We use MoP trained on AGNews using Llama3-8B-Instruct as in Section~\ref{sec:exp_steer}; then, during the simulation, we replace Llama3-8B-Instruct with Gemma2-9B-Instruct~\citep{team2024gemma} and Mistralv0.3-7B-Instruct~\citep{jiang2023mistral}. As reported in Table~\ref{tab:transfer}, MoP can be effectively transferred to other models and even enjoys the improvement over MoP with Llama3-8B-Instruct.

\begin{table}[t]
    \centering
    \caption{Transferability of MoP on AGNews dataset. Note that we use instruction-finetuned version for all language models.}
    \small
    \begin{tabular}{cccc}
        \toprule
         & FID$\downarrow$ & MAUVE$\uparrow$ & KL Cosine$\downarrow$ \\ \midrule
        MoP Llama3-8B   & 0.951 & 0.871 & 0.069   \\
        \hdashline
        $\rightarrow$ Gemma2-9B   & 0.492 & 0.957 & 0.006   \\ 
        $\rightarrow$ Mistral-7B  & 0.923 & 0.869 & 0.081 \\
        \bottomrule
        \end{tabular}
    \label{tab:transfer}
    \vspace{-5mm}
\end{table}

\begin{table}[ht]
    \small
    \centering
    \caption{Ablation study: MoP without exemplars or Persona Synthesizer.}
    \begin{tabular}{cccc}
        \toprule
         & FID$\downarrow$ & MAUVE$\uparrow$ & KL Cosine$\downarrow$ \\ \midrule
         MoP & 0.951 & 0.871 & 0.069   \\
        \hdashline
         w/o exemplars  & 3.694 & 0.552 & 0.560   \\ 
         w/o persona syn & 1.674 & 0.807 & 0.174   \\
         w random personas & 1.814 & 0.622 & 0.061   \\
         \bottomrule
        \end{tabular}
    \label{tab:ablation_misc}
\end{table}

\subsection{Ablation Studies}
We conduct extensive ablation studies to examine different variants of the proposed MoP methods. Unless otherwise specified, the experimental settings are consistent with those used in the steerability experiment described in Section~\ref{sec:exp_steer}.

\noindent\textbf{Component Ablations.} To investigate the effect of each component, we ablate exemplars and the persona synthesizer from MoP described in Section~\ref{sec:exemplar_mop}. For MoP without the persona synthesizer, we will use 100 predefined personas generated from GPT-4 without any context on the dataset. For MoP with random personas, we randomly select from 100 synthesized personas and exemplars are chosen heuristically based on the highest embedding similarity to the chosen persona embedding. The results in Table~\ref{tab:ablation_misc} indicate that prompting with exemplars is crucial to steer LLM responses toward the population of interest. Meanwhile, the persona synthesizer helps significantly improve the alignment score.

\noindent\textbf{Ablation on Varying the Number of Personas and Exemplars.} We investigate the impact of the number of personas and exemplars used for MoP. Specifically, we vary the number of personas in the range $\{20, 50, 100, 200\}$ and the number of exemplars in the range $\{100, 200, 1000, 2000, 5000\}$. The result in Figure~\ref{fig:abl_vary} shows that generally, more fine-grained persona descriptions and more exemplars are typically helpful for MoP. However, the performance is saturated at 2000 exemplars.

\noindent\textbf{Mixing Personas in Single Query.} In previous experiments, we randomly choose a single persona for inference. In this experiment, we explore the possibility of prompting LLMs with multiple personas in a single query. Note that, unlike the training, mixing personas is not a trivial problem if we want to maintain the black-box usage of LLM during inference. Thus, we prompt the pretrained LLM to synthesize a new persona description given $L$ persona-exemplars sampled according to the mixing weights given by the gating network. The mixed persona will be used to construct a new prompt with $L$ exemplars. The prompt is given in Appendix~\ref{sec:prompts}. The results in Table~\ref{tab:abla_top_m} demonstrate that combining multiple personas can enhance the diversity of generated responses and, in certain cases (e.g., $L=2$), improve alignment. However, using a larger number of personas ($L = 4, 8$) may negatively impact response alignment.

\begin{table}[t]
    \small
    \centering
    \caption{The performance of MoP on AGNews when mixing different personas during inference.}
    \begin{tabular}{cccc}
        \toprule
         & FID$\downarrow$ & MAUVE$\uparrow$ & KL Cosine$\downarrow$ \\ \midrule
        MoP no mixing  & 0.951 & 0.871 & 0.069   \\ 
        \hdashline
        MoP w/ $L = 2$   & \textbf{0.776} & \textbf{0.925} & \textbf{0.039}   \\ 
        MoP w/ $L = 4$    & 0.988 & 0.898 & 0.059   \\ 
        MoP w/ $L = 8$   & 1.152 & 0.870 & 0.062  \\ \hline 
        \end{tabular}
    \label{tab:abla_top_m}
    \vspace{-3mm}
\end{table}

\section{Related Work}
\noindent\textbf{LLM Steerability.} Recent work has explored understanding and directing the opinions of LLMs~\citep{santurkar2023whose, li2023steerability, scherrer2024evaluating, sorensen2024roadmap}. For example,~\citet{santurkar2023whose} introduced OpinionQA to evaluate alignment with 60 U.S. demographic groups, revealing notable discrepancies. \citet{scherrer2024evaluating} examined moral beliefs encoded in LLMs using moral scenario surveys. Methods to enhance LLM steerability include prompt engineering and finetuning~\citep{feng2023pretraining, kim2024few, zhao2023group, hwang2023aligning, simmons2022moral}. \citet{santurkar2023whose} designed prompts to target specific demographics. FERMI~\citep{kim2024few} uses personalized prompts and in-context examples for user-specific outputs, while GPO~\citep{zhao2023group} aligns LLMs with group opinion distributions through finetuning.

\noindent\textbf{Zeroshot Synthetic Data Generation.} Generating synthetic datasets with LLMs has been widely studied~\citep{zou2024fusegen, gupta2023targen, yu2024large, gao2022self, ye2022progen, yu2023regen}. Progen~\citep{ye2022progen} uses noise-robust influence functions to guide high-quality sample generation. FuseGen~\citep{zou2024fusegen} builds on this by employing multiple LMs for generating synthetic samples but requires access to classifier models. In contrast, our MoP method avoids dependency on classifier families and multi-round processes, while also serving as a plugin to improve diversity in Progen's framework.

\noindent\textbf{Sampling Diversity.} LLMs often exhibit low sampling diversity, leading to lexical redundancy even when feedback-tuned or persona-prompted~\citep{santurkar2023whose, padmakumar2023does}. Padmakumar et al.\cite{padmakumar2023does} found that InstructGPT reduced diversity compared to human-generated text. Efforts to balance quality and diversity, such as GAN-based approaches\citep{xu2018dp, caccia2018language}, often fail to outperform LLMs~\citep{holtzman2019curious}. Techniques like temperature tuning~\citep{hinton2015distilling, guo2017calibration} and dynamic sampling~\citep{zhang2024edt} aim to improve diversity, though recent work questions their efficacy~\citep{peeperkorn2024temperature, renze2024effect}. Diversity is critical in applications requiring human-representative outputs, such as synthetic data for testing social theories or scaling experimental research~\citep{horton2023large, aher2023using, argyle2023out}. However, LLMs often fail to represent human behavior accurately, exhibiting biases and inconsistencies~\citep{santurkar2023whose, dorner2023personality, aher2023using}. These limitations highlight challenges in modeling individual or group actions.

\section{Conclusion}
We introduced Mixture of Personas (MoP), a probabilistic prompting method designed to align LLM-generated responses with a target population. Our experiments showed that exemplar-based variants of MoP significantly improve the alignment and diversity of synthetic data, leading to enhanced performance in downstream tasks. Furthermore, MoP is flexible and does not require fine-tuning of the base model, enabling seamless transferability across different models without retraining.

\section{Limitations}
While MoP is a prompting method and can be transferable to different models, MoP still needs access to LLM output logits to be trainable. This might be a constraint for an application of MoP to closed-source models such as ChatGPT, etc. However, this constraint is less restrictive compared to other finetuning baselines or requiring access to the persona datasets. Furthermore, as our study focuses on task scenarios where access to personal data is restricted due to privacy concerns, we recognize the potential trade-offs between protecting privacy and avoiding subjective labeling of users. By designing persona prompts without personal data, there is a risk of introducing biases inherent in the construction of these prompts, which may not fully represent user populations. Addressing this limitation will require deeper investigations into the interplay between privacy and fairness, such as exploring techniques for bias mitigation within the probabilistic prompting framework. We leave this investigation for future work.

\bibliography{refs}

\appendix

\section{Societal Impact}
While generating human-like datasets opens up many exciting possibilities for augmenting blind spots in data that may have otherwise led to limitations preventing a complete understanding of important information, synthetic data can still lack specific nuances, diversity, and edge cases present in human-derived data. This is especially prevalent in population sub-groups that remain underrepresented, causing a lack of labeled data in specific languages that affect the authenticity and coverage of over two billion individuals worldwide \citep{joshi2020state}. Our focus on augmenting human records instead of replacing them is notwithstanding the potential of synthetic data; it is imperative that we continue supporting data collection obtained directly from people in under-served communities and domains and, whenever possible, make those data freely available for empirical and experimental research. Beyond the behavioral science and data generation for training ML models, data generated using MoP could be leveraged across a broad range of applications, such as informatics-based health claims, market intelligence and user-preference modeling, and free-form text writing – where the goal is not only to produce accurate or factually-correct answers but also to simulate a population of interest otherwise out of reach.

\section{Additional Experiment Settings} \label{sec:add_exp_setting}

\textbf{Datasets.} Below are the detailed descriptions of four datasets used in this paper.

\begin{itemize}[leftmargin=5mm]
    \item \textit{AGNews}~\cite{zhang2015character}: This dataset contains 103600 news articles crawled from AG News. Each article is categorized into political news, sports news, business news, and technology news.
    \item \textit{Yelp Reviews}~\cite{zhang2015character}: This dataset contains restaurant reviews from the Yelp platform. Each review is labeled with a positive or negative sentiment label. This popular dataset is used for sentiment analysis and opinion mining.
    \item \textit{SST-2 (Stanford Sentiment Treebank)}~\cite{socher-etal-2013-recursive}: The dataset consists of 11,855 individual sentences extracted from movie reviews to analyze the compositional effects of sentiment in language. Each sentence is annotated with a binary label indicating whether the review is positive or negative.
    \item \textit{IMDB Reviews}~\cite{maas-etal-2011-learning}: The dataset consists of 50,000 movie reviews crawled from IMDB with 25,000 samples for positive classes and 25,000 samples for negative classes.
\end{itemize}

\textbf{Baselines.} We provide descriptions for baselines used in our paper.
\begin{itemize}[leftmargin=5mm]
    \item \textbf{ZeroGen}~\cite{ye2022zerogen} generates synthetic samples by feeding simple class-conditioned prompts to language models. We reuse the same prompts described in~\cite{yu2024large}.
    \item \textbf{Attrprompt}~\cite{yu2024large} further increases the diversity of synthetic samples by utilizing a set of sample attributes generated by chat-GPT. For example, to generate synthetic examples for AgNews, Attrprompt also inputs randomized values for `writing style' or `location' to the prompt in addition to class information. We refer the readers to their papers for the full description of the used attributes for each dataset. In our experiment, we use the same attribute set released by the authors in their repository~\footnote{https://github.com/yueyu1030/AttrPrompt/tree/main}.
    \item \textbf{ProGen~\citep{ye2022progen}:} proposes a framework for zero-shot dataset generation that iteratively improves the quality of synthetic datasets. It uses feedback from a task-specific model, guided by a noise-tolerant influence function, to identify high-quality samples and incorporate them as in-context examples for subsequent data generation.
    \item \textbf{PICLE~\citep{choi2024picle}:} is a framework designed to elicit specific target personas from large language models (LLMs) using an in-context learning (ICL) approach. It employs a novel likelihood-ratio-based selection mechanism to identify and prepend the most informative examples from a persona-specific pool, guiding the model to align with the desired persona. By leveraging Bayesian inference, PICLe modifies the LLM's output distribution to better reflect the target persona, achieving improved persona elicitation compared to baseline methods.
\end{itemize}  

\textbf{Metrics.}
Measuring the alignment of the synthetic data with the golden dataset can be challenging due to the discrete nature of text sequences. We follow~\citep{pillutla2021mauve, yu2024large} to evaluate the alignment and diversity of the simulated data in the embedding space. Throughout, we use a pre-trained sentence encoder `all-mpnet-base-v2'~\footnote{sentence-transformers/all-mpnet-base-v2}~\cite{song2020mpnet} as a text encoder to map generated sentences into a common vector space. We then report three following metrics:
\begin{itemize}[leftmargin=5mm]
    \item FID (Fréchet Inception Distance)~\cite{heusel2017gans} is a widely-used metric in computer vision literature to measure the fidelity of synthetic images with respect to real images. Following the approach in ~\citep{yue2022synthetic}, we adapt FID for text generation. We first calculate the mean and covariance matrices of sentence embeddings for both the golden test set and synthetic datasets, then use the Fréchet distance to compute the FID between them.

    \item The MAUVE metric~\cite{pillutla2021mauve, pillutla2023mauve} evaluates the similarity between two text distributions by comparing their divergence frontiers. It begins by embedding text into vectors and clustering them to create histograms of cluster assignments. A divergence curve is then constructed from these histograms, and the area under the curve quantifies the difference between the synthetic and golden data distributions. In our experiments, we use the original implementation of MAUVE~\footnote{https://krishnap25.github.io/mauve/}. As suggested by the authors, we set the number of clusters to 500 and the scaling parameter to 1. 
    
    \item The Kullback-Leibler Cosine (KL Cosine) metric evaluates the divergence between the pairwise cosine similarity distributions of a method and the ground truth. First, the pairwise cosine similarities are computed for the method and the ground truth. Then, the Kullback-Leibler divergence between their normalized histograms is calculated as:
    \begin{equation}
        D_{KL}(P \| Q) = \sum_i P(i) \log \frac{P(i)}{Q(i)}   
    \end{equation}
where \(P(i)\) represents the histogram of pairwise cosine similarities for the method, and \(Q(i)\) represents the histogram for the ground truth.
\end{itemize}

\begin{table}[t]
    \centering
    \caption{Train test split and the number of classes in each dataset.}
    \resizebox{\linewidth}{!}{
    \begin{tabular}{ccccc}
        \toprule
         & AgNews & Yelp & SST-2 & IMDB \\ \midrule
        Train set   & 96000 & 141076 & 67349 & 14006\\ 
        Test set   & 7600 & 35323 & 1821 & 14056\\ 
        Number of classes   & 4  & 2  & 2  & 2\\ \bottomrule 
    \end{tabular}
    }
    \label{tab:dataset_statistics}
\end{table}

\section{Additional ablation studies}
\label{sec:ablation_study}
We conducted an additional experiment to evaluate whether our framework is sensitive to the format of ICL prompts. Specifically, we modified the original ICL prompt (`You have written the following news blurb: [example]') to two alternative formats: ICL Template 1 and ICL Template 2. The results in the table below indicate that changes in the ICL format have minimal impact on performance in our setting.

\begin{verbatim}
ICL Template 1: 
Example news blurb: [example]

ICL Template 2: 
The following news blurb 
is given as an example: [example]
\end{verbatim}

These results suggest that our framework’s performance is not overly sensitive to the specific format of ICL prompts. 

\begin{table}[h]
    \centering
    \caption{Comparison of metrics across different ICL prompts and templates.}
    \resizebox{\linewidth}{!}{
    \begin{tabular}{lcccc}
        \toprule
        & FID & Mauve & Cosine Similarity & KL Cosine \\ \midrule
        Original ICL Prompt & 0.951 & 0.871 & 0.108 & 0.069 \\
        ICL Template 1      & 0.955 & 0.883 & 0.110 & 0.074 \\
        ICL Template 2      & 0.955 & 0.881 & 0.108 & 0.069 \\ \bottomrule
    \end{tabular}
    }
    \label{tab:icl_metrics}
\end{table}

\section{MoP Prompts}
\label{sec:prompts}

In this section, we present the prompts used throughout our experiments. 
\begin{enumerate}
    \item \textbf{Synthesizing New Samples.} The prompts in Fig.~\ref{fig:agnews_mop_prompt}, Fig.~\ref{fig:yelp_mop_prompt}, Fig.~\ref{fig:sst2_mop_prompt}, and Fig.~\ref{fig:imdb_mop_prompt} are designed to synthesize new samples based on provided examples. These prompts guide the language model to generate new samples in line with the provided context and example, while maintaining consistency with the style and tone of the given input.
    \item \textbf{Generating new personas.} We use the prompts in Fig.~\ref{fig:agnews_mop_gen_prompt}, Fig.~\ref{fig:yelp_mop_gen_prompt}, Fig.~\ref{fig:sst2_mop_gen_prompt}, and Fig.~\ref{fig:imdb_mop_gen_prompt} to generate a new persona based on examples that reflect the characteristics of that persona.
    \item \textbf{Classifying personas.} We use the prompts shown in Fig.~\ref{fig:agnews_mop_classify_prompt} to classify a persona based on its description. The prompt provides a persona description of a reporter, and the task is to identify the reporter's specialization from four predefined categories: world news, sports news, business news, or sci/tech news.
    \item \textbf{Persona Mixing Prompt.} The prompt in Fig.~\ref{fig:agnews_mop_mix_prompt} is designed to synthesize a unified persona description by combining multiple persona descriptions with sample news blurbs written by the reporter. It integrates the thematic focus, stylistic tendencies, and primary interests from the inputs to generate a concise and cohesive profile of the reporter.
\end{enumerate}

\begin{figure*}
\caption{MoP Prompt for Agnews} 
\label{fig:agnews_mop_prompt}
\lstset{
  basicstyle=\ttfamily\small, 
  frame=single,              
  breaklines=true,           
  breakatwhitespace=true,    
  columns=fullflexible       
}
\begin{lstlisting}
<|begin_of_text|><|start_header_id|>system<|end_header_id|>\n\nYou embody the persona in the description to complete the tasks.<|eot_id|><|start_header_id|>user<|end_header_id|>\n\n
\n\nYou have written the following news blurb: {example}
\n\nPlease write a short news blurb similar to the above blurb.<|eot_id|> <|start_header_id|>assistant<|end_header_id|>\n\n
\end{lstlisting}
\end{figure*}

\begin{figure*}
\caption{MoP Prompt for Yelp}
\label{fig:yelp_mop_prompt}
\lstset{
  basicstyle=\ttfamily\small, 
  frame=single,              
  breaklines=true,           
  breakatwhitespace=true,    
  columns=fullflexible       
}
\begin{lstlisting}
<|begin_of_text|><|start_header_id|>system<|end_header_id|>\n\nYou embody the persona in the description to complete the tasks.<|eot_id|><|start_header_id|>user<|end_header_id|>\n\n
\n\nYour review for the restaurant {context}: {example}
\n\nPlease write a short review for the restaurant {context}, similar to the above review:<|eot_id|> <|start_header_id|>assistant<|end_header_id|>\n\n
\end{lstlisting}
\end{figure*}

\begin{figure*}
\caption{MoP Prompt for SST-2}
\label{fig:sst2_mop_prompt}
\lstset{
  basicstyle=\ttfamily\small, 
  frame=single,              
  breaklines=true,           
  breakatwhitespace=true,    
  columns=fullflexible       
}
\begin{lstlisting}
<|begin_of_text|><|start_header_id|>system<|end_header_id|>\n\nYou embody the persona in the description to complete the tasks.<|eot_id|><|start_header_id|>user<|end_header_id|>\n\n
\n\nYou have written the following review: {example}
\n\nPlease write a review sentence, similar to the above review:<|eot_id|><|start_header_id|>assistant<|end_header_id|> \n\n<|start_header_id|>assistant<|end_header_id|>\n\n
\end{lstlisting}
\end{figure*}

\begin{figure*}
\caption{MoP Prompt for IMDB}
\label{fig:imdb_mop_prompt}
\lstset{
  basicstyle=\ttfamily\small, 
  frame=single,              
  breaklines=true,           
  breakatwhitespace=true,    
  columns=fullflexible       
}
\begin{lstlisting}
<|begin_of_text|><|start_header_id|>system<|end_header_id|>\n\nYou embody the persona in the description to complete the tasks.<|eot_id|><|start_header_id|>user<|end_header_id|>\n\n
\n\nYou have written the following review for the movie {context}: {example}
\n\nPlease write a review for the movie {context}, similar to the above review:<|eot_id|><|start_header_id|>assistant<|end_header_id|>\n\n
\end{lstlisting}
\end{figure*}

\begin{figure*}
\caption{MoP Prompt to generate persona for Agnews}
\label{fig:agnews_mop_gen_prompt}
\lstset{
  basicstyle=\ttfamily\small, 
  frame=single,              
  breaklines=true,           
  breakatwhitespace=true,    
  columns=fullflexible       
}
\begin{lstlisting}
<|begin_of_text|><|start_header_id|>system<|end_header_id|>

You are a helpful AI assistant<|eot_id|><|start_header_id|>user<|end_header_id|>

Given a list of news blurbs written by a reporter, construct a concise persona description that reflects the reporter's primary focus, style, and thematic interests.

Example persona descriptions:
You are a sports reporter, specializing in baseball news, with a focus on the Major League Baseball (MLB) playoffs and postseason games.

List of news blurbs:
{examples}

Generate a short persona description that synthesizes their focus, preferences, and stylistic tendencies into a single cohesive statement.<|eot_id|><|start_header_id|>assistant<|end_header_id|>

The short persona is:
\end{lstlisting}
\end{figure*}

\begin{figure*}
\caption{MoP Prompt to generate persona for Yelp}
\label{fig:yelp_mop_gen_prompt}
\lstset{
  basicstyle=\ttfamily\small, 
  frame=single,              
  breaklines=true,           
  breakatwhitespace=true,    
  columns=fullflexible       
}
\begin{lstlisting}
<|begin_of_text|><|start_header_id|>system<|end_header_id|>

You are a helpful AI assistant<|eot_id|><|start_header_id|>user<|end_header_id|>

Given a list of restaurant review written by a customer, construct a concise persona description that reflects the customer's preferences, writing style, and interests.

Examples of Persona Descriptions:
You are a food critic, specializing in fine dining and gourmet cuisine with a focus on presentation and taste.
You are a casual diner who enjoys comfort food and writes personal and informal reviews.

List of reviews:
{examples}

Generate a short persona description that synthesizes their interests, preferences, and writing stylistic tendencies into a single cohesive statement.<|eot_id|><|start_header_id|>assistant<|end_header_id|>

The short persona is:
\end{lstlisting}
\end{figure*}

\begin{figure*}
\caption{MoP Prompt to generate persona for SST-2}
\label{fig:sst2_mop_gen_prompt}
\lstset{
  basicstyle=\ttfamily\small, 
  frame=single,              
  breaklines=true,           
  breakatwhitespace=true,    
  columns=fullflexible       
}
\begin{lstlisting}
<|begin_of_text|><|start_header_id|>system<|end_header_id|>

You are a helpful AI assistant<|eot_id|><|start_header_id|>user<|end_header_id|>

Given a list of movie review written by a viewer, construct a concise persona description that reflects the review's preferences, writing style, and thematic interests.

Examples of Persona Descriptions:
You are a movie critic, specializing in horror movies and independent films with a focus on cinematography and storytelling.
You are a casual viewer who enjoys action-packed films and writes personal and informal reviews.

List of reviews:
{examples}

Generate a short persona description that synthesizes their focus, preferences, and stylistic tendencies into a single cohesive statement.<|eot_id|><|start_header_id|>assistant<|end_header_id|>

The short persona is:
\end{lstlisting}
\end{figure*}

\begin{figure*}
\caption{MoP Prompt to generate persona for IMDB} \label{fig:imdb_mop_gen_prompt}
\lstset{
  basicstyle=\ttfamily\small, 
  frame=single,              
  breaklines=true,           
  breakatwhitespace=true,    
  columns=fullflexible       
}
\begin{lstlisting}
<|begin_of_text|><|start_header_id|>system<|end_header_id|>

You are a helpful AI assistant<|eot_id|><|start_header_id|>user<|end_header_id|>

Given a list of movie review written by a viewer, construct a concise persona description that reflects the review's preferences, writing style, and thematic interests.

Examples of Persona Descriptions:
You are a movie critic, specializing in horror movies and independent films with a focus on cinematography and storytelling.
You are a casual viewer who enjoys action-packed films and writes personal and informal reviews.

List of reviews:
{examples}

Generate a short persona description that synthesizes their focus, preferences, and stylistic tendencies into a single cohesive statement.<|eot_id|><|start_header_id|>assistant<|end_header_id|>

The short persona is:
\end{lstlisting}
\end{figure*}

\begin{figure*}
\caption{MoP Prompt to classify persona for AgNews}
\label{fig:agnews_mop_classify_prompt}
\lstset{
  basicstyle=\ttfamily\small, 
  frame=single,              
  breaklines=true,           
  breakatwhitespace=true,    
  columns=fullflexible       
}
\begin{lstlisting}
<|begin_of_text|><|start_header_id|>system<|end_header_id|>

You are a helpful AI assistant<|eot_id|><|start_header_id|>user<|end_header_id|>

Given a persona description of a reporter, choose the specialization of the personas: world news, sports news, business news, sci/tech news.

Persona description: You are a technology reporter, specializing in the intersection of hardware and software, with a focus on the PC industry, Linux, and major tech companies like Microsoft, Apple, and IBM.
Options:
A. world news
B. sports news
C. business news
D. sci/tech news
Answer: D

Persona description: {}
Options:
A. world news
B. sports news
C. business news
D. sci/tech news
Answer: <|eot_id|><|start_header_id|>assistant<|end_header_id|>
\end{lstlisting}
\end{figure*}

\begin{figure*}
\caption{MoP Prompt to mix personas for AgNews}
\label{fig:agnews_mop_mix_prompt}
\lstset{
  basicstyle=\ttfamily\small, 
  frame=single,              
  breaklines=true,           
  breakatwhitespace=true,    
  columns=fullflexible       
}
\begin{lstlisting}
<|begin_of_text|><|start_header_id|>system<|end_header_id|>

You are a highly capable and insightful AI assistant<|eot_id|><|start_header_id|>user<|end_header_id|>

Given a list of persona descriptions of a reporter and a list of news blurbs he/she has written, construct a concise persona description that reflects the reporter's primary focus, style, and thematic interests.

Example persona descriptions:
You are a sports reporter, specializing in baseball news, with a focus on the Major League Baseball (MLB) playoffs and postseason games.

Inputs:
1. List of persona descriptions: A list of general persona characteristics previously associated with the reporter.
2. List of news blurbs: A selection of sample news blurbs authored by the reporter, which reveal their tone, thematic focus, and writing style.

Using the inputs, generate a short persona description that synthesizes their focus, preferences, and stylistic tendencies into a single cohesive statement.

List of persona descriptions:
{personas}

List of news blurbs:
{examples}

Please provide the short persona description.<|eot_id|><|start_header_id|>assistant<|end_header_id|>

The short persona is:
\end{lstlisting}
\end{figure*}

\end{document}

%% file: style.tex
\usepackage{graphicx}
\usepackage{float,epstopdf}
\usepackage{bbm}

\usepackage{microtype}

\usepackage{natbib}

\usepackage{subcaption}
\usepackage{booktabs}

\usepackage{tcolorbox}
\usepackage{amsmath}
\usepackage{mathtools}
\usepackage{amsthm}
\usepackage{dsfont}
\usepackage{multicol}
\usepackage{multirow} 
\usepackage{amsfonts} 
\usepackage{mathrsfs}
\usepackage{fancyhdr}
\usepackage[amssymb, thickqspace]{SIunits}
\usepackage{enumitem}
\usepackage{pgfplotstable}
\usepackage{arydshln}
\usepackage{lipsum}		
\usepackage{hyperref}
\usepackage{fancyvrb}
\usepackage{cases}
\usepackage{verbatimbox}
\usepackage{listings} 
\usepackage{xcolor}
\usepackage[ruled, linesnumbered]{algorithm2e}

\usepackage{url}

\usepackage{thmtools}
\usepackage{thm-restate}
\usepackage{tabu}
\usepackage{adjustbox}

%% file: math_commands.tex
\newcommand{\mc}{\mathcal}

\newcommand{\mbf}{\mathbf}

\def\eqref#1{equation~(\ref{#1})}

\def\1{\bm{1}}

\DeclareMathAlphabet{\mathsfit}{\encodingdefault}{\sfdefault}{m}{sl}
\SetMathAlphabet{\mathsfit}{bold}{\encodingdefault}{\sfdefault}{bx}{n}

\newcommand{\R}{\mathbb{R}}

\newcommand{\softmax}{\mathrm{softmax}}




%% file: main.bbl
\begin{thebibliography}{56}
\providecommand{\natexlab}[1]{#1}

\bibitem[{Aher et~al.(2023)Aher, Arriaga, and Kalai}]{aher2023using}
Gati~V Aher, Rosa~I Arriaga, and Adam~Tauman Kalai. 2023.
\newblock Using large language models to simulate multiple humans and replicate human subject studies.
\newblock In \emph{International Conference on Machine Learning}, pages 337--371. PMLR.

\bibitem[{Argyle et~al.(2023)Argyle, Busby, Fulda, Gubler, Rytting, and Wingate}]{argyle2023out}
Lisa~P Argyle, Ethan~C Busby, Nancy Fulda, Joshua~R Gubler, Christopher Rytting, and David Wingate. 2023.
\newblock Out of one, many: Using language models to simulate human samples.
\newblock \emph{Political Analysis}, 31(3):337--351.

\bibitem[{Bisbee et~al.(2023)Bisbee, Clinton, Dorff, Kenkel, and Larson}]{bisbee2023synthetic}
James Bisbee, Joshua~D Clinton, Cassy Dorff, Brenton Kenkel, and Jennifer~M Larson. 2023.
\newblock Synthetic replacements for human survey data? the perils of large language models.
\newblock \emph{Political Analysis}, pages 1--16.

\bibitem[{Brown(2020)}]{brown2020language}
Tom~B Brown. 2020.
\newblock Language models are few-shot learners.
\newblock \emph{arXiv preprint arXiv:2005.14165}.

\bibitem[{Caccia et~al.(2018)Caccia, Caccia, Fedus, Larochelle, Pineau, and Charlin}]{caccia2018language}
Massimo Caccia, Lucas Caccia, William Fedus, Hugo Larochelle, Joelle Pineau, and Laurent Charlin. 2018.
\newblock Language {GAN}s falling short.
\newblock \emph{arXiv preprint arXiv:1811.02549}.

\bibitem[{Chang et~al.(2023)Chang, Reitter, Aksitov, and Sung}]{chang2023kl}
Chung-Ching Chang, David Reitter, Renat Aksitov, and Yun-Hsuan Sung. 2023.
\newblock Kl-divergence guided temperature sampling.
\newblock \emph{arXiv preprint arXiv:2306.01286}.

\bibitem[{Chen et~al.(2024)Chen, Liu, Huang, Wu, Liu, Jiang, Pu, Lei, Chen, Wang et~al.}]{chen2024large}
Jin Chen, Zheng Liu, Xu~Huang, Chenwang Wu, Qi~Liu, Gangwei Jiang, Yuanhao Pu, Yuxuan Lei, Xiaolong Chen, Xingmei Wang, et~al. 2024.
\newblock When large language models meet personalization: Perspectives of challenges and opportunities.
\newblock \emph{World Wide Web}, 27(4):42.

\bibitem[{Choi and Li(2024)}]{choi2024picle}
Hyeong~Kyu Choi and Yixuan Li. 2024.
\newblock {PICLe}: Eliciting diverse behaviors from large language models with persona in-context learning.
\newblock In \emph{Forty-first International Conference on Machine Learning}.

\bibitem[{Dorner et~al.(2023)Dorner, S{\"u}hr, Samadi, and Kelava}]{dorner2023personality}
Florian~E Dorner, Tom S{\"u}hr, Samira Samadi, and Augustin Kelava. 2023.
\newblock Do personality tests generalize to large language models?
\newblock \emph{arXiv preprint arXiv:2311.05297}.

\bibitem[{Dubey et~al.(2024)Dubey, Jauhri, Pandey, Kadian, Al-Dahle, Letman, Mathur, Schelten, Yang, Fan et~al.}]{dubey2024llama}
Abhimanyu Dubey, Abhinav Jauhri, Abhinav Pandey, Abhishek Kadian, Ahmad Al-Dahle, Aiesha Letman, Akhil Mathur, Alan Schelten, Amy Yang, Angela Fan, et~al. 2024.
\newblock The llama 3 herd of models.
\newblock \emph{arXiv preprint arXiv:2407.21783}.

\bibitem[{Feng et~al.(2023)Feng, Park, Liu, and Tsvetkov}]{feng2023pretraining}
Shangbin Feng, Chan~Young Park, Yuhan Liu, and Yulia Tsvetkov. 2023.
\newblock From pretraining data to language models to downstream tasks: Tracking the trails of political biases leading to unfair nlp models.
\newblock \emph{arXiv preprint arXiv:2305.08283}.

\bibitem[{Gao et~al.(2022)Gao, Pi, Lin, Xu, Ye, Wu, Zhang, Liang, Li, and Kong}]{gao2022self}
Jiahui Gao, Renjie Pi, Yong Lin, Hang Xu, Jiacheng Ye, Zhiyong Wu, Weizhong Zhang, Xiaodan Liang, Zhenguo Li, and Lingpeng Kong. 2022.
\newblock Self-guided noise-free data generation for efficient zero-shot learning.
\newblock \emph{arXiv preprint arXiv:2205.12679}.

\bibitem[{Guo et~al.(2017)Guo, Pleiss, Sun, and Weinberger}]{guo2017calibration}
Chuan Guo, Geoff Pleiss, Yu~Sun, and Kilian~Q Weinberger. 2017.
\newblock On calibration of modern neural networks.
\newblock In \emph{International conference on machine learning}, pages 1321--1330. PMLR.

\bibitem[{Gupta et~al.(2023)Gupta, Scaria, Anantheswaran, Verma, Parmar, Sawant, Mishra, and Baral}]{gupta2023targen}
Himanshu Gupta, Kevin Scaria, Ujjwala Anantheswaran, Shreyas Verma, Mihir Parmar, Saurabh~Arjun Sawant, Swaroop Mishra, and Chitta Baral. 2023.
\newblock Targen: Targeted data generation with large language models.
\newblock \emph{arXiv preprint arXiv:2310.17876}.

\bibitem[{Heusel et~al.(2017)Heusel, Ramsauer, Unterthiner, Nessler, and Hochreiter}]{heusel2017gans}
Martin Heusel, Hubert Ramsauer, Thomas Unterthiner, Bernhard Nessler, and Sepp Hochreiter. 2017.
\newblock Gans trained by a two time-scale update rule converge to a local nash equilibrium.
\newblock \emph{Advances in neural information processing systems}, 30.

\bibitem[{Hinton(2015)}]{hinton2015distilling}
Geoffrey Hinton. 2015.
\newblock Distilling the knowledge in a neural network.
\newblock \emph{arXiv preprint arXiv:1503.02531}.

\bibitem[{Holtzman et~al.(2019)Holtzman, Buys, Du, Forbes, and Choi}]{holtzman2019curious}
Ari Holtzman, Jan Buys, Li~Du, Maxwell Forbes, and Yejin Choi. 2019.
\newblock The curious case of neural text degeneration.
\newblock \emph{arXiv preprint arXiv:1904.09751}.

\bibitem[{Horton(2023)}]{horton2023large}
John~J Horton. 2023.
\newblock Large language models as simulated economic agents: What can we learn from homo silicus?
\newblock Technical report, National Bureau of Economic Research.

\bibitem[{Hwang et~al.(2023)Hwang, Majumder, and Tandon}]{hwang2023aligning}
EunJeong Hwang, Bodhisattwa Majumder, and Niket Tandon. 2023.
\newblock \href {https://doi.org/10.18653/v1/2023.findings-emnlp.393} {Aligning language models to user opinions}.
\newblock In \emph{Findings of the Association for Computational Linguistics: EMNLP 2023}, pages 5906--5919, Singapore. Association for Computational Linguistics.

\bibitem[{Jiang et~al.(2023)Jiang, Sablayrolles, Mensch, Bamford, Chaplot, de~las Casas, Bressand, Lengyel, Lample, Saulnier et~al.}]{jiang2023mistral}
AQ~Jiang, A~Sablayrolles, A~Mensch, C~Bamford, DS~Chaplot, D~de~las Casas, F~Bressand, G~Lengyel, G~Lample, L~Saulnier, et~al. 2023.
\newblock Mistral 7b (2023).
\newblock \emph{arXiv preprint arXiv:2310.06825}.

\bibitem[{Jordan and Jacobs(1994)}]{jordan1994hierarchical}
Michael~I Jordan and Robert~A Jacobs. 1994.
\newblock Hierarchical mixtures of experts and the em algorithm.
\newblock \emph{Neural computation}, 6(2):181--214.

\bibitem[{Joshi et~al.(2020)Joshi, Santy, Budhiraja, Bali, and Choudhury}]{joshi2020state}
Pratik Joshi, Sebastin Santy, Amar Budhiraja, Kalika Bali, and Monojit Choudhury. 2020.
\newblock The state and fate of linguistic diversity and inclusion in the {NLP} world.
\newblock \emph{arXiv preprint arXiv:2004.09095}.

\bibitem[{Kim and Yang(2024)}]{kim2024few}
Jaehyung Kim and Yiming Yang. 2024.
\newblock Few-shot personalization of llms with mis-aligned responses.
\newblock \emph{arXiv preprint arXiv:2406.18678}.

\bibitem[{Li et~al.(2023{\natexlab{a}})Li, Mehrabi, Peris, Goyal, Chang, Galstyan, Zemel, and Gupta}]{li2023steerability}
Junyi Li, Ninareh Mehrabi, Charith Peris, Palash Goyal, Kai-Wei Chang, Aram Galstyan, Richard Zemel, and Rahul Gupta. 2023{\natexlab{a}}.
\newblock On the steerability of large language models toward data-driven personas.
\newblock \emph{arXiv preprint arXiv:2311.04978}.

\bibitem[{Li et~al.(2023{\natexlab{b}})Li, Zhu, Lu, and Yin}]{li2023synthetic}
Zhuoyan Li, Hangxiao Zhu, Zhuoran Lu, and Ming Yin. 2023{\natexlab{b}}.
\newblock Synthetic data generation with large language models for text classification: Potential and limitations.
\newblock \emph{arXiv preprint arXiv:2310.07849}.

\bibitem[{Maas et~al.(2011)Maas, Daly, Pham, Huang, Ng, and Potts}]{maas-etal-2011-learning}
Andrew~L. Maas, Raymond~E. Daly, Peter~T. Pham, Dan Huang, Andrew~Y. Ng, and Christopher Potts. 2011.
\newblock \href {https://aclanthology.org/P11-1015} {Learning word vectors for sentiment analysis}.
\newblock In \emph{Proceedings of the 49th Annual Meeting of the Association for Computational Linguistics: Human Language Technologies}, pages 142--150, Portland, Oregon, USA. Association for Computational Linguistics.

\bibitem[{Mallapragada et~al.(2010)Mallapragada, Jin, and Jain}]{mallapragada2010non}
Pavan~Kumar Mallapragada, Rong Jin, and Anil Jain. 2010.
\newblock Non-parametric mixture models for clustering.
\newblock In \emph{Joint IAPR International Workshops on Statistical Techniques in Pattern Recognition (SPR) and Structural and Syntactic Pattern Recognition (SSPR)}, pages 334--343. Springer.

\bibitem[{Padmakumar and He(2023)}]{padmakumar2023does}
Vishakh Padmakumar and He~He. 2023.
\newblock Does writing with language models reduce content diversity?
\newblock \emph{arXiv preprint arXiv:2309.05196}.

\bibitem[{Peeperkorn et~al.(2024)Peeperkorn, Kouwenhoven, Brown, and Jordanous}]{peeperkorn2024temperature}
Max Peeperkorn, Tom Kouwenhoven, Dan Brown, and Anna Jordanous. 2024.
\newblock Is temperature the creativity parameter of large language models?
\newblock \emph{arXiv preprint arXiv:2405.00492}.

\bibitem[{Pillutla et~al.(2023)Pillutla, Liu, Thickstun, Welleck, Swayamdipta, Zellers, Oh, Choi, and Harchaoui}]{pillutla2023mauve}
Krishna Pillutla, Lang Liu, John Thickstun, Sean Welleck, Swabha Swayamdipta, Rowan Zellers, Sewoong Oh, Yejin Choi, and Zaid Harchaoui. 2023.
\newblock Mauve scores for generative models: Theory and practice.
\newblock \emph{Journal of Machine Learning Research}, 24(356):1--92.

\bibitem[{Pillutla et~al.(2021)Pillutla, Swayamdipta, Zellers, Thickstun, Welleck, Choi, and Harchaoui}]{pillutla2021mauve}
Krishna Pillutla, Swabha Swayamdipta, Rowan Zellers, John Thickstun, Sean Welleck, Yejin Choi, and Zaid Harchaoui. 2021.
\newblock Mauve: Measuring the gap between neural text and human text using divergence frontiers.
\newblock \emph{Advances in Neural Information Processing Systems}, 34:4816--4828.

\bibitem[{Renze and Guven(2024)}]{renze2024effect}
Matthew Renze and Erhan Guven. 2024.
\newblock The effect of sampling temperature on problem solving in large language models.
\newblock \emph{arXiv preprint arXiv:2402.05201}.

\bibitem[{Sanh(2019)}]{sanh2019distilbert}
V~Sanh. 2019.
\newblock Distilbert, a distilled version of bert: Smaller, faster, cheaper and lighter.
\newblock \emph{arXiv preprint arXiv:1910.01108}.

\bibitem[{Santurkar et~al.(2023)Santurkar, Durmus, Ladhak, Lee, Liang, and Hashimoto}]{santurkar2023whose}
Shibani Santurkar, Esin Durmus, Faisal Ladhak, Cinoo Lee, Percy Liang, and Tatsunori Hashimoto. 2023.
\newblock Whose opinions do language models reflect?
\newblock In \emph{International Conference on Machine Learning}, pages 29971--30004. PMLR.

\bibitem[{Scherrer et~al.(2024)Scherrer, Shi, Feder, and Blei}]{scherrer2024evaluating}
Nino Scherrer, Claudia Shi, Amir Feder, and David Blei. 2024.
\newblock Evaluating the moral beliefs encoded in llms.
\newblock \emph{Advances in Neural Information Processing Systems}, 36.

\bibitem[{Shazeer et~al.(2017)Shazeer, Mirhoseini, Maziarz, Davis, Le, Hinton, and Dean}]{shazeer2017outrageously}
Noam Shazeer, Azalia Mirhoseini, Krzysztof Maziarz, Andy Davis, Quoc Le, Geoffrey Hinton, and Jeff Dean. 2017.
\newblock Outrageously large neural networks: The sparsely-gated mixture-of-experts layer.
\newblock \emph{arXiv preprint arXiv:1701.06538}.

\bibitem[{Simmons(2022)}]{simmons2022moral}
Gabriel Simmons. 2022.
\newblock Moral mimicry: Large language models produce moral rationalizations tailored to political identity. arxiv.
\newblock \emph{arXiv preprint arXiv:2209.12106}.

\bibitem[{Socher et~al.(2013)Socher, Perelygin, Wu, Chuang, Manning, Ng, and Potts}]{socher-etal-2013-recursive}
Richard Socher, Alex Perelygin, Jean Wu, Jason Chuang, Christopher~D. Manning, Andrew Ng, and Christopher Potts. 2013.
\newblock \href {https://aclanthology.org/D13-1170} {Recursive deep models for semantic compositionality over a sentiment treebank}.
\newblock In \emph{Proceedings of the 2013 Conference on Empirical Methods in Natural Language Processing}, pages 1631--1642, Seattle, Washington, USA. Association for Computational Linguistics.

\bibitem[{Song et~al.(2020)Song, Tan, Qin, Lu, and Liu}]{song2020mpnet}
Kaitao Song, Xu~Tan, Tao Qin, Jianfeng Lu, and Tie-Yan Liu. 2020.
\newblock Mpnet: Masked and permuted pre-training for language understanding.
\newblock \emph{Advances in neural information processing systems}, 33:16857--16867.

\bibitem[{Sorensen et~al.(2024)Sorensen, Moore, Fisher, Gordon, Mireshghallah, Rytting, Ye, Jiang, Lu, Dziri et~al.}]{sorensen2024roadmap}
Taylor Sorensen, Jared Moore, Jillian Fisher, Mitchell Gordon, Niloofar Mireshghallah, Christopher~Michael Rytting, Andre Ye, Liwei Jiang, Ximing Lu, Nouha Dziri, et~al. 2024.
\newblock A roadmap to pluralistic alignment.
\newblock \emph{arXiv preprint arXiv:2402.05070}.

\bibitem[{Sun et~al.(2024)Sun, Yang, Reddy, Fung, Chan, Zhai, and Ji}]{sun2024persona}
Chenkai Sun, Ke~Yang, Revanth~Gangi Reddy, Yi~R Fung, Hou~Pong Chan, ChengXiang Zhai, and Heng Ji. 2024.
\newblock Persona-db: Efficient large language model personalization for response prediction with collaborative data refinement.
\newblock \emph{arXiv preprint arXiv:2402.11060}.

\bibitem[{Team et~al.(2024)Team, Riviere, Pathak, Sessa, Hardin, Bhupatiraju, Hussenot, Mesnard, Shahriari, Ram{\'e} et~al.}]{team2024gemma}
Gemma Team, Morgane Riviere, Shreya Pathak, Pier~Giuseppe Sessa, Cassidy Hardin, Surya Bhupatiraju, L{\'e}onard Hussenot, Thomas Mesnard, Bobak Shahriari, Alexandre Ram{\'e}, et~al. 2024.
\newblock Gemma 2: Improving open language models at a practical size.
\newblock \emph{arXiv preprint arXiv:2408.00118}.

\bibitem[{T{\"o}rnberg(2023)}]{tornberg2023chatgpt}
Petter T{\"o}rnberg. 2023.
\newblock Chatgpt-4 outperforms experts and crowd workers in annotating political twitter messages with zero-shot learning.
\newblock \emph{arXiv preprint arXiv:2304.06588}.

\bibitem[{Tseng et~al.(2024)Tseng, Huang, Hsiao, Hsu, Foo, Huang, and Chen}]{tseng2024two}
Yu-Min Tseng, Yu-Chao Huang, Teng-Yun Hsiao, Yu-Ching Hsu, Jia-Yin Foo, Chao-Wei Huang, and Yun-Nung Chen. 2024.
\newblock Two tales of persona in llms: A survey of role-playing and personalization.
\newblock \emph{arXiv preprint arXiv:2406.01171}.

\bibitem[{Wang et~al.(2023)Wang, Mao, Wu, Ge, Wei, and Ji}]{wang2023unleashing}
Zhenhailong Wang, Shaoguang Mao, Wenshan Wu, Tao Ge, Furu Wei, and Heng Ji. 2023.
\newblock Unleashing the emergent cognitive synergy in large language models: A task-solving agent through multi-persona self-collaboration.
\newblock \emph{arXiv preprint arXiv:2307.05300}.

\bibitem[{Xu et~al.(2018)Xu, Ren, Lin, and Sun}]{xu2018dp}
Jingjing Xu, Xuancheng Ren, Junyang Lin, and Xu~Sun. 2018.
\newblock {DP-GAN}: diversity-promoting generative adversarial network for generating informative and diversified text.
\newblock \emph{arXiv preprint arXiv:1802.01345}.

\bibitem[{Ye et~al.(2022{\natexlab{a}})Ye, Gao, Feng, Wu, Yu, and Kong}]{ye2022progen}
Jiacheng Ye, Jiahui Gao, Jiangtao Feng, Zhiyong Wu, Tao Yu, and Lingpeng Kong. 2022{\natexlab{a}}.
\newblock Progen: Progressive zero-shot dataset generation via in-context feedback.
\newblock \emph{arXiv preprint arXiv:2210.12329}.

\bibitem[{Ye et~al.(2022{\natexlab{b}})Ye, Gao, Li, Xu, Feng, Wu, Yu, and Kong}]{ye2022zerogen}
Jiacheng Ye, Jiahui Gao, Qintong Li, Hang Xu, Jiangtao Feng, Zhiyong Wu, Tao Yu, and Lingpeng Kong. 2022{\natexlab{b}}.
\newblock Zerogen: Efficient zero-shot learning via dataset generation.
\newblock \emph{arXiv preprint arXiv:2202.07922}.

\bibitem[{Yu et~al.(2024{\natexlab{a}})Yu, Luo, Wei, Lei, Huang, Hao, and Zhu}]{yu2024neeko}
Xiaoyan Yu, Tongxu Luo, Yifan Wei, Fangyu Lei, Yiming Huang, Peng Hao, and Liehuang Zhu. 2024{\natexlab{a}}.
\newblock Neeko: Leveraging dynamic lora for efficient multi-character role-playing agent.
\newblock \emph{arXiv preprint arXiv:2402.13717}.

\bibitem[{Yu et~al.(2024{\natexlab{b}})Yu, Zhuang, Zhang, Meng, Ratner, Krishna, Shen, and Zhang}]{yu2024large}
Yue Yu, Yuchen Zhuang, Jieyu Zhang, Yu~Meng, Alexander~J Ratner, Ranjay Krishna, Jiaming Shen, and Chao Zhang. 2024{\natexlab{b}}.
\newblock Large language model as attributed training data generator: A tale of diversity and bias.
\newblock \emph{Advances in Neural Information Processing Systems}, 36.

\bibitem[{Yu et~al.(2023)Yu, Zhuang, Zhang, Meng, Shen, and Zhang}]{yu2023regen}
Yue Yu, Yuchen Zhuang, Rongzhi Zhang, Yu~Meng, Jiaming Shen, and Chao Zhang. 2023.
\newblock Regen: Zero-shot text classification via training data generation with progressive dense retrieval.
\newblock \emph{arXiv preprint arXiv:2305.10703}.

\bibitem[{Yue et~al.(2022)Yue, Inan, Li, Kumar, McAnallen, Shajari, Sun, Levitan, and Sim}]{yue2022synthetic}
Xiang Yue, Huseyin~A Inan, Xuechen Li, Girish Kumar, Julia McAnallen, Hoda Shajari, Huan Sun, David Levitan, and Robert Sim. 2022.
\newblock Synthetic text generation with differential privacy: A simple and practical recipe.
\newblock \emph{arXiv preprint arXiv:2210.14348}.

\bibitem[{Zhang et~al.(2024)Zhang, Bao, and Huang}]{zhang2024edt}
Shimao Zhang, Yu~Bao, and Shujian Huang. 2024.
\newblock {EDT}: Improving large language models' generation by entropy-based dynamic temperature sampling.
\newblock \emph{arXiv preprint arXiv:2403.14541}.

\bibitem[{Zhang et~al.(2015)Zhang, Zhao, and LeCun}]{zhang2015character}
Xiang Zhang, Junbo Zhao, and Yann LeCun. 2015.
\newblock Character-level convolutional networks for text classification.
\newblock \emph{Advances in neural information processing systems}, 28.

\bibitem[{Zhao et~al.(2023)Zhao, Dang, and Grover}]{zhao2023group}
Siyan Zhao, John Dang, and Aditya Grover. 2023.
\newblock Group preference optimization: Few-shot alignment of large language models.
\newblock \emph{arXiv preprint arXiv:2310.11523}.

\bibitem[{Zou et~al.(2024)Zou, Liu, Li, Zhang, Liu, and Zhang}]{zou2024fusegen}
Tianyuan Zou, Yang Liu, Peng Li, Jianqing Zhang, Jingjing Liu, and Ya-Qin Zhang. 2024.
\newblock Fusegen: Plm fusion for data-generation based zero-shot learning.
\newblock \emph{arXiv preprint arXiv:2406.12527}.

\end{thebibliography}
